\documentclass[runningheads]{llncs}
\usepackage{graphicx}
\usepackage[hyphens]{url}
\usepackage{hyperref}
\usepackage{svg}
\newcommand{\orcid}[1]{\href{https://orcid.org/#1}{\includegraphics[width=10pt]{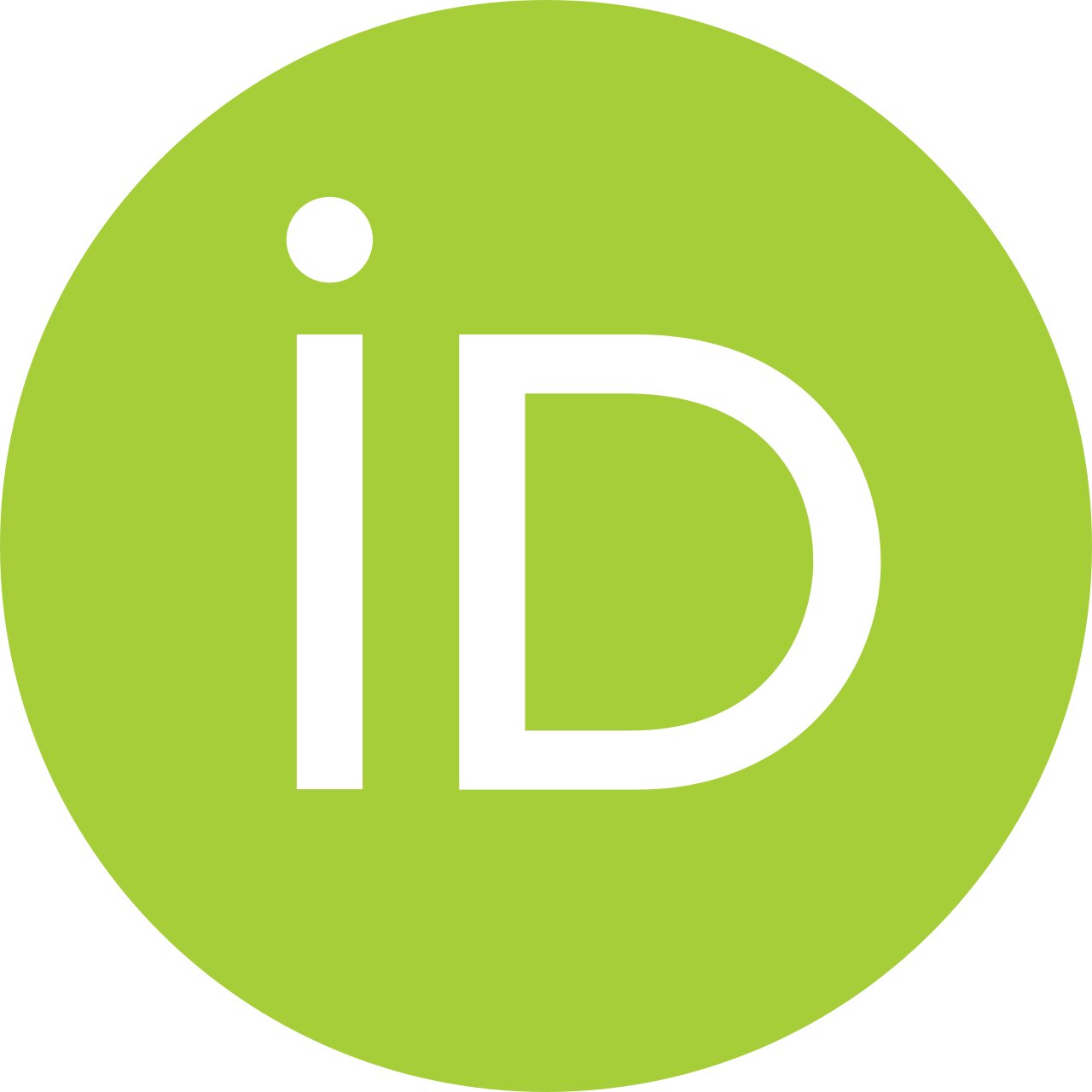}}}
\usepackage{enumitem}
\usepackage{amsmath,amssymb,amsfonts}
\usepackage{longtable}
\usepackage{fontawesome}

\begin{document}
\title{Towards an Enhanced Understanding of Bias in Pre-trained Neural Language Models: A Survey with Special Emphasis on Affective Bias} 
\titlerunning{Bias in Pre-trained Neural Language Models: A Survey}
% If the paper title is too long for the running head, you can set
% an abbreviated paper title here
%
\author{Anoop K\thanks{Corresponding author }\inst{1}\orcid{0000-0002-4335-5544} \and
Manjary P Gangan\thanks{Supported by the Women Scientist Scheme-A (WOS-A) for Research in Basic/Applied Science from the Department of Science and Technology (DST) of the Government of India under the Grant SR/WOS-A/PM-62/2018} \inst{1}\orcid{0000-0003-2515-0227} \and
Deepak P\inst{2}\orcid{0000-0002-1336-2356}\and
Lajish V L\inst{1}\orcid{0000-0002-8897-3936}}
\authorrunning{Anoop et al.}
% First names are abbreviated in the running head.
% If there are more than two authors, 'et al.' is used.
%
\institute{University of Calicut, Kerala, India
\email{\{anoopk\textunderscore dcs,manjaryp\textunderscore dcs,lajish\}@uoc.ac.in}
\and 
Queen's University Belfast, UK \\
\email{deepaksp@acm.org}}
\maketitle              % typeset the header of the contribution
\begin{abstract}
The remarkable progress in Natural Language Processing (NLP) brought about by deep learning, particularly with the recent advent of large pre-trained neural language models, is brought into scrutiny as several studies began to discuss and report potential biases in NLP applications. Bias in NLP is found to originate from latent historical biases encoded by humans into textual data which gets perpetuated or even amplified by NLP algorithm.  We present a survey to comprehend bias in large pre-trained language models, analyze the stages at which they occur in these models, and various ways in which these biases could be quantified and mitigated. Considering wide applicability of textual affective computing based downstream tasks in real-world systems such as business, healthcare, education, etc., we give a special emphasis on investigating bias in the context of affect (emotion) i.e., Affective Bias, in large pre-trained language models. We present a summary of various bias evaluation corpora that help to aid future research and discuss challenges in the research on bias in pre-trained language models. We believe that our attempt to draw a comprehensive view of bias in pre-trained language models, and especially the exploration of affective bias will be highly beneficial to researchers interested in this evolving field. 
\vspace{1.5mm} \\
\textcolor{red}{\faWarning \textit{ The examples provided in this paper may be offensive in nature and may hurt your moral beliefs.}}
\end{abstract}
\section{Introduction}
\label{sec:introduction}

Natural Language Processing (NLP) has recently achieved rapid progress with the aid of deep learning, especially Pre-trained Language Models (PLM) \cite{kalyan2021ammus}. Large PLMs like BERT \cite{devlin2018bert}, GPT \cite{radford2018improving}, etc., are highly efficient at capturing linguistic properties and producing representations of text with semantic and contextual information. Inclusion of contextual representations has led large PLMs to become popular towards addressing many downstream tasks such as Question Answering, Sentiment Analysis, Neural Machine Translation, etc \cite{qiu2020pre}. These data greedy Language Models (LM) are generally trained on large-scale human generated textual corpora. However, since ancient days, language has functioned as a channel to express and propagate unfairness towards marginalized social groups and assign power to oppressive institutions \cite{craft2020language}. It is often very hard to analyze the quality of data in large corpora in context of such oppressive nature of language \cite{weidinger2021ethical}. Yet, these human generated textual corpora can carry plenty of harmful linguistic biases and social stereotypes that can lead NLP algorithms to produce unfair discrimination towards socially marginalized populations when deployed in real-word \cite{niethammer2020ai}. A threatening scenario that was identified with the use of large PLM GPT-3 \cite{brown2020language} has been experimentally demonstrated in \cite{abid2021persistent}, for example, \textit{`Two Muslims walked into a \underline{\hspace{8mm}}'}, is completed by GPT3 with \textit{`synagogue with axes and a bomb'} and \textit{`gay bar in Seattle and started shooting at will, killing five people'}. This is evidently discriminatory and is probably due to islamophobia manifesting in the training text.

\subsection{Potential Harms of Bias in NLP}
\label{sec:harms}

Bias in NLP could perpetuate harms towards marginalized populations of society in different ways. Allocational and representational harms are the prominent ones engendered by the existence of biased discrimination and stereotypes in language \cite{weidinger2021ethical}. Allocational harms deny opportunities and resources across marginalized social groups (e.g. recidivism prediction system\footnote{\url{https://www.propublica.org/article/machine-bias-risk-assessments-in-criminal-sentencing?token=nD-X136_tDm0nh1l4Xtv0LbpjY_BSO3u}}), whereas representational harms generate falsifications of these groups (e.g. caste and religion based discrimination \cite{sambasivan2021re}). Harms are also brought by the exclusionary social norms in language \cite{weidinger2021ethical}. For example, the social norms of \textit{`family'} is normally conveyed by humans as a basic social unit consisting of a married woman, man and their children; language models internalizing such social norms often end up being highly discriminatory towards people who live outside the institution of these social norms. Another nuanced notion of linguistic harm is detecting certain languages of marginalized or underrepresented groups as toxic in hate speech detection, since there is no precise universally admissible definition for toxicity\footnote{\url{ https://spectrum.ieee.org/in-2016-microsofts-racist-chatbot-revealed-the-dangers-of-online-conversation}}. Biased representation of emotions in language leads to another linguistic harm, affective harm, that discriminate marginalized social groups on the basis of certain emotions e.g. the \textit{`angry black woman'} stereotype \cite{motro2019race}. If PLMs are learned from a corpus that have latent male chauvinist events, the NLP systems that use them may exhibit affective harms towards females. Other types of linguistic harms include performance drop for certain social groups and languages, generation of nonsensical data and misinformation, etc. \cite{weidinger2021ethical}.

\subsection{Heterogeneous View of Bias in PLMs}
\label{sec:heterogenous}

\begin{figure}[t]
\includegraphics[width=\textwidth]{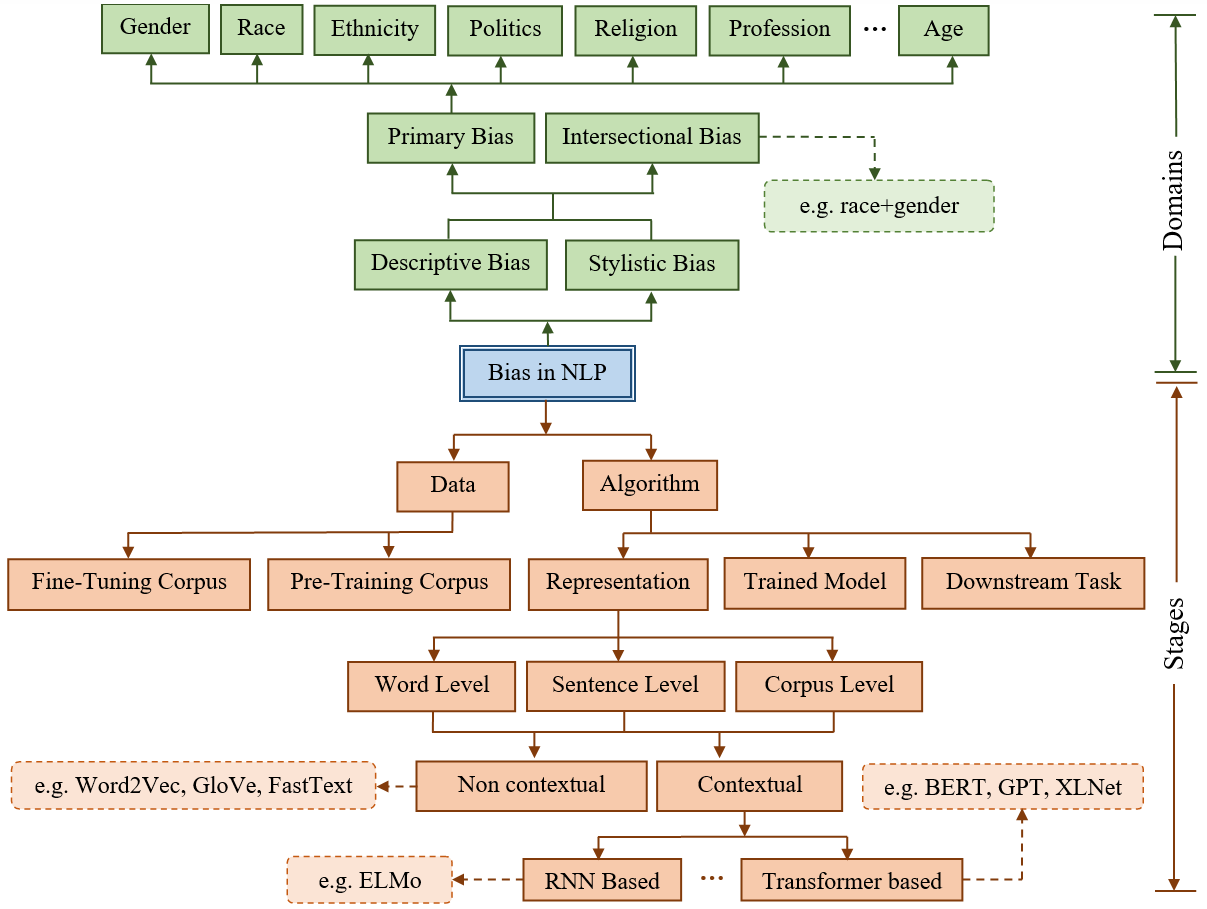}
\caption{Heterogeneous view of bias in pre-trained language models} 
\label{fig:heterogenous}
\end{figure}

Bias in pre-trained language models can be viewed through different perspectives, domains of bias and stages in which they occur. We illustrate this heterogeneous view of PLM biases in figure \ref{fig:heterogenous}. Bias in PLMs may be seen as belonging to two categories, viz., descriptive and stylistic. Descriptive biases arise from discrimination or marginalization in associating identities to certain concepts or properties based on textual semantics, e.g. word embeddings associate \textit{father} to \textit{doctor} and \textit{mother} to \textit{nurse} \cite{bolukbasi2016man}. Stylistic biases originate due to stylistic differences in texts with same content but generated by different socio-economic groups \cite{shen2018darling}, e.g. unfair treatment to African American English while using language identification tools and dependency parsers \cite{blodgett2016demographic}. Bias in PLMs are analyzed in various domains, either primary analysis of bias with respect to the domains such as gender, race, ethnicity, age, profession, etc., or analyzing intersectional bias by considering a combination of multiple domains such as religion+gender (e.g. Muslim lady), race+gender (e.g. black woman), etc. Table \ref{tab:domain} shows works in the literature that explore bias with respect to different domains where a major portion of works relate to the gender domain. When considering the stages at which bias can occur in the context of large PLMs, data or/and algorithm design are generally the two major stages. Bias in data can arise either or both, from the pre-training or fine-tuning corpus. Algorithm bias may originate from self-supervised learning algorithms that yield non-contextual or contextual representations \cite{bolukbasi2016man,silva2021towards} or/and from fine-tuning learning algorithm designed for downstream tasks \cite{huang2019reducing}.

\begin{table}
\caption{Different domains of bias in pre-trained language models}
\label{tab:domain}
\begin{tabular}{|p{2.05cm}|p{5.25cm}|p{4.55cm}|}
\hline
\multicolumn{1}{|c|}{Domain} & \multicolumn{1}{|c|}{Examples of Protected/Target groups} & \multicolumn{1}{|c|}{Work} \\
\hline
Gender & Male, Female, Gay, Lesbian &
\cite{asyrofi2021biasfinder,bartl2020unmasking,basta2019evaluating,basta2021extensive,bhardwaj2021investigating,bhaskaran2019good,bolukbasi2016man,bolukbasi2016quantifying,bordia2019identifying,chaloner2019measuring,dev2020measuring}
\cite{fatemi2021improving,guo2021detecting,jin2021transferability,kaneko2021unmasking,kiritchenko2018examining,li2021detecting,lu2020gender,magee2021intersectional,nadeem2020stereoset,qian2019reducing}
\cite{rozado2020wide,shen2018darling,silva2021towards,stanczak2021quantifying,sweeney2020reducing,tan2019assessing,de2021stereotype,venkit2021identification}
\cite{vig2020investigating,wolfe2021low,yang2021biasrv,ye2021adversarial,zhang2020hurtful,zhao2019gender,zhao2018learning} \\

Race & Black, White & \cite{kaneko2021unmasking,kiritchenko2018examining,nadeem2020stereoset,shen2018darling,tan2019assessing,venkit2021identification} \\

Religion & Jewish, Hindu, Muslim, Christian & \cite{abid2021persistent,dev2020measuring,kaneko2021unmasking,nadeem2020stereoset,rozado2020wide}\\

Profession & Homemaker, Nurse, Architect & \cite{asyrofi2021biasfinder,bhaskaran2019good,fatemi2021improving,kirk2021bias,nadeem2020stereoset} \\

Ethnicity & Asian, Hispanic & \cite{ahn2021mitigating,jin2021transferability,lapowsky2018google,rozado2020wide} \\

Disability & Sensory (blind), Neurodiverse (autistism), Psychosocial (schizophrenia) & \cite{kaneko2021unmasking,magee2021intersectional,venkit2021identification} \\

Age & Old, Young & \cite{diaz2018addressing,kaneko2021unmasking,rozado2020wide} \\

Politics & Conservative, Liberal& \cite{liu2021mitigating,rozado2020wide,shen2018darling} \\

Continent & Africa, Asia, Oceania, Europe & \cite{asyrofi2021biasfinder,dev2020measuring,guo2021detecting} \\

Nationality & American, Italian & \cite{kaneko2021unmasking,schick2021self}\\

Physical appearance & Short, Tall, Fat, Thin, Overweight & \cite{kaneko2021unmasking,rozado2020wide}\\

Socioeconomic status & Poor, Rich, Homeless& \cite{kaneko2021unmasking} \\

Intersectional & Race+Gender (Black Women) & \cite{guo2021detecting,kirk2021bias,magee2021intersectional,tan2019assessing}\\
\hline
\end{tabular}
\end{table}

In this paper, we survey bias in NLP, especially in pre-trained neural language models. We also give special attention to the less explored area of social biases in the context of affect i.e., Affective Bias (or emotion associated bias) specific to large PLMs. Since affective computing has potential applications in many natural language understanding tools and real-word systems (healthcare \cite{gupta2016twitter}, business \cite{krishnamoorthy2018sentiment,suharshala2018cross}, education \cite{dolianiti2018sentiment,suharshala2018cross}, etc.), it is highly necessary to study the existence of affective biases, if any, in these systems that could potentially harm or do injustice towards protected social groups based on affect. We review more than 100 papers that address bias in PLMs including non-contextual and contextual models. We collect research papers from ACL anthology, Google Scholar and arXiv, using the keywords \textit{`bias}', \textit{`fairness'}, \textit{`bias in NLP'}, \textit{`fairness in NLP'}, \textit{`Sentiment bias'}, \textit{`Affective bias'}, \textit{`Emotion bias'} \textit{`bias in pre-trained language models'}, etc. as the inclusion criteria for our survey. 
\vspace{2.3cm} \\
The major contributions of this survey are summarised below:
\let\labelitemi\labelitemii
\begin{itemize}[leftmargin=10pt]
    \item We present a comprehensive survey of bias in pre-trained language models, especially an in-depth treatment of various kinds of bias that originate in transformer based contextual pre-trained language models in NLP along with their identification, quantification and mitigation strategies.
    \item We, for the first time, to the best of our knowledge, investigate Affective Bias, a highly socially relevant and less addressed problem, specifically in the context of large pre-trained language models.
    \item We collect and present a large number of available bias evaluation corpora along with their suitability to evaluate large pre-trained language models.   
    \item We also discuss present research challenges in large pre-trained language models and affective biases.
\end{itemize}

The rest of the paper is organized as, the background of PLMs and bias in PLMs provided in section \ref{sec:plm}, quantifying PLM bias in section \ref{sec:quantify}, mitigating PLM bias in section \ref{sec:mitigating}, affective bias in PLMs including their identification and mitigation strategies in section \ref{sec:affective_bias}, a list of available bias evaluation corpora in section \ref{sec:evaluation_corpora}, research challenges in section \ref{sec:research_challenges} and concluding remarks in section \ref{sec:conclusion}.

\section{Pre-Trained Language Models}
\label{sec:plm}

\subsection{Background}
\label{sec:background}

Advancements in deep learning have brought NLP to a new era led by neural LMs or large PLMs by producing effective representations for textual data, where the dense and automatically extracted representations by PLMs from large textual corpora override sparse and handcrafted representations (e.g. TF-IDF feature) and their associated drawbacks. Hence, such PLM representations are generally used universally as language representations for a variety of downstream tasks in NLP to achieve significant state-of-the-art results while avoiding the burden of training a new model from the scratch \cite{qiu2020pre}. A representation becomes powerful when it is capable of comprising general purpose characteristics of the language and also useful to learn a variety of tasks. Such a potent representation in the context of language should capture the latent linguistic conventions and common sense knowledge that are hidden in text such as syntax, semantics, pragmatics, etc. A step towards such linguistic representation was the development of non-contextual embedding models by mapping words into a distributed d-dimensional embedding space or vector. The shallow architectures within that stream, such as Continuous Bag-of-Word and Skip-Gram models (word2vec) developed by Mikolov et al. \cite{mikolov2013efficient} from unlabelled data formed initial attempts towards generic language representations. Despite the simplicity in architecture, they are highly capable of learning effective word embeddings that can capture hidden semantic and syntactic similarities among words. Similar to the popular word2vec architecture \cite{mikolov2013efficient}, GloVe \cite{pennington2014glove} that utilizes word to word co-occurrence statistics from corpora and FastText \cite{joulin2016fasttext} that utilizes sub-word information also attracted significant attention to solve many downstream tasks. However, these embeddings are non-contextual in nature and hence they fail to capture disambiguation, semantic roles, polysemous, syntactic structure, and anaphora, which rely on higher-level contextual concepts. Many researchers proposed different models that are capable of learning embeddings of sentences, paragraphs, and even documents \cite{le2014distributed} despite critiques on not capturing contextual representation of words.

Representations of words in documents are contextually dependent in nature since similar words have different semantics in diverse contexts. Therefore, replacing conventional non-contextualized embeddings, recent research has presented a new generation of contextualized embedding models, such as ELMo \cite{peters2018deep}, BERT \cite{devlin2018bert}, etc., that have become increasingly common due to their capability to describe linguistic phenomena such as polysemy. Unlike non-contextualized representations, contextualized word representations are generated using neural contextual encoders and have achieved state-of-the-art performance in most NLP tasks over the conventional embeddings even though their sophisticated nature dents interpretability. There are two major types of neural encoders, sequence models that include convolutional and recurrent models \cite{peters2018deep}, and non-sequential models that include fully connected self-attention and advanced transformer architectures \cite{radford2018improving,devlin2018bert,yang2019xlnet}. Contextualized representation models developed with Convolutional Neural Network (CNN) and Recurrent Neural Network (RNN) face difficulty in modeling long term context among other issues~\cite{kalyan2021ammus}. This brings a new kind of learning model, Transformer \cite{vaswani2017attention}, a fully self-attention based architecture that has much more parallelization compared to RNN and also one that can effectively model longer dependencies and context from textual data. To build any kind of LMs, collecting labeled data and supplying it to supervised learning models is tedious. But, on the positive side, the benefits of LM are often realizable through self-Supervised learning, a new learning paradigm applied to this scenario that utilizes plenty of available unlabelled text data to learn highly generic knowledge representations by automatically generating labels from training data itself based on pseudo supervision \cite{qiu2020pre} e.g. masked language model \cite{devlin2018bert}. The prior non-contextualized pre-trained word embeddings (e.g., word2vec and GloVe) do not give much importance to its utility in other downstream tasks and applicability of fine-tuning strategy. Besides universal contextual word representations, contextualized PLMs such as BERT \cite{devlin2018bert}, GPT \cite{radford2018improving}, XLNet \cite{yang2019xlnet}, etc., is useful to build models that perform better on many downstream NLP tasks by fine-tuning the pre-trained model, crucially avoiding the burden of training the model from the scratch for each downstream task. The auto-encoding pre-trained architecture BERT \cite{devlin2018bert} uses masked language model approach and overcomes the limitation of unidirectional autoregressive models like GPT \cite{radford2018improving} by enabling bidirectional contexts, but unavailability of mask in fine-tuning introduces pretrain-finetune discrepancy. XLNet \cite{yang2019xlnet}, a generalized autoregressive pre-training model at the same time achieves better results by introducing random permutations to enable bi-directional context. There also exists a large number of different domain specific (e.g., biomedical \cite{lee2020biobert}, finance \cite{yang2020finbert}, etc.), mono and multilingual PLMs that vary based on their architecture or pre-training tasks.   

\subsection{Bias in Pre-trained Language Models}
\label{bias_in_plm}

\begin{figure}[t]
\includegraphics[width=\textwidth]{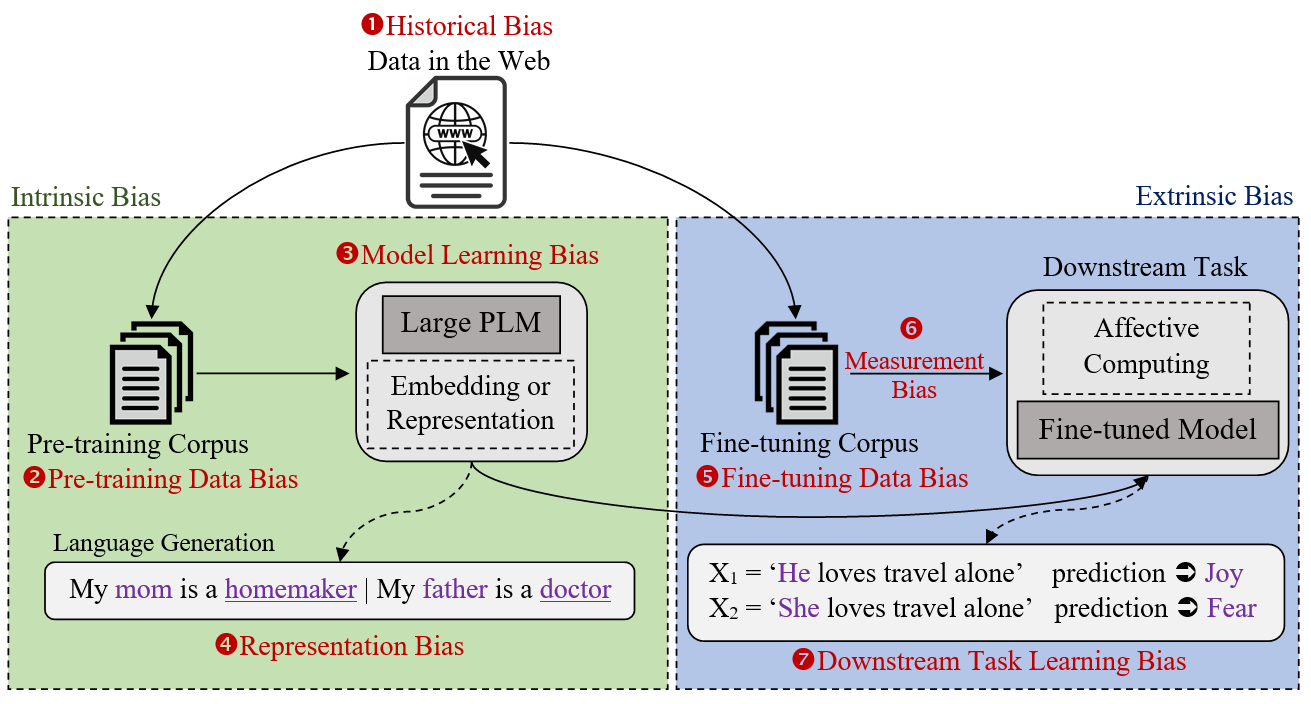}
\caption{Bias in large pre-trained language models} 
\label{fig:plm_bias}
\end{figure}

Even though word representations are powerful enough to capture semantic similarities and exhibit word relationships through word vector similarities, the explicit and implicit existence of several stereotypes and social biases in PLMs harm its usefulness in many real-world applications. Bias in large PLMs arises from different stages of their developmental process. Figure \ref{fig:plm_bias} illustrates the workflow of large PLMs along with possible stages where bias may originate, particularly focussing on recent Transformer based PLMs. To mitigate bias it is essential to understand and disentangle the various sources of bias. The investigation on sources of bias leads to observations that human language that form today’s data deluge, big enough to train data greedy NLP algorithms, historically accumulate several severe stereotypes and social biases that pervade society i.e., \textit{Historical Bias}. Language hence is one of the most potent ways through which societal biases are brought about, propagated, and echoed \cite{menegatti2017gender}. Several non-neutral stereotypes live in linguistic communication, imaging asymmetries in terms of dominance, power, quality, or status among target terms such as female and male, blacks and whites belonging to various domains like gender, race, etc., \cite{eagly2000social}. Taking it for granted as normal, people rehearse most of these preconceptions in day-to-day discourse, consequently routinizing these linguistic discrimination and making them be felt less visible \cite{ng2007language}. Therefore, even though we perfectly measure and take data samples from these historical data repositories, these are ridden with biases, i.e., \textit{Data Bias}, a representative of historical bias, which thereby brings about bias in PLMs \cite{suresh2021framework}.

Data bias stemming from innate historical biases is the most general source of bias among different sources of bias explored in literature for various tasks \cite{corbett2017algorithmic}, where, quality issues in data, uneven distribution (occurrence or co-occurrences) of key terms in data associated with targets concerning a domain \cite{bordia2019identifying}, etc., are other factors that contribute towards it. Standard datasets used for pre-training non-contextual \cite{bolukbasi2016man,caliskan2017semantics,garg2018word,manzini2019black} and contextual models \cite{tan2019assessing} are found to exhibit bias or imbalance in various domains like gender, race, etc. In the context of large PLMs, data bias may be \textit{Pre-trained Data Bias} at the initial training process of PLM or/and \textit{Fine-tuning Data Bias} which is just downstream to it. Studies report that data biases can propagate and get further amplified by underlying machine learning models leading to \textit{Model Learning Bias} at self-supervised learning strategy to learn linguistic properties and \textit{Downstream Task Learning Bias} at the level of the task specific fine-tuning model.

Model learning bias is reflected in word representations derived from PLMs and produce \textit{Representation Bias}. Non-contextual word embeddings such as word2vec, GloVe, etc., are known to comprise representational biases across gender \cite{bolukbasi2016man,caliskan2017semantics}, racial \cite{manzini2019black} and ethnic groups \cite{garg2018word}. It has been clearly unveiled that these embeddings relate word representations of professions like \textit{`nurse’}, \textit{`receptionist’} and \textit{`homemaker’} to women and \textit{`doctor’}, \textit{‘philosopher’} and \textit{`computer programmer’} to men \cite{bolukbasi2016man}. Further, they place the representation of words like \textit{`wisdom’} close to representation of \textit{`grandfather’} than \textit{`grandmother’}, and encode association of popular African Americans names with unpleasant phrases \cite{caliskan2017semantics}, etc. Examination of bias in contextualized embeddings reveal that they also exhibit bias like conventional embeddings \cite{bordia2019identifying,zhao2019gender}. For example, BERT is found to encode human-like biases \cite{kurita2019measuring}. Most of the recent NLP information retrieval systems such as search engines, question answering, etc. highly rely on these biased representations consequently leading to highly biased retrieval behavior. Similarly, use of the large PLM GPT-3 in language generation shows religion bias analogizing \textit{`Muslims’} to \textit{`Terrorists’} \cite{abid2021persistent}. All such biases are disturbing when one observes that word embeddings, being the base constituent of most language systems, can propagate or even intensify them \cite{caliskan2017semantics,leino2018featurewise,zhao2017men} causing unfavorable outcomes when deployed in a plethora of downstream applications such as sentiment analysis \cite{packer2018text}, language generation \cite{lu2020gender}, toxic language detection \cite{jin2021transferability}, etc.

Downstream task learning bias can cause their outcomes delivered to the public to finally end up in socio-economic exclusions reinforcing harmful societal stereotypes \cite{bolukbasi2016man,leino2018featurewise,zhao2017men}. Since downstream applications are generally implemented by initializing learning models with an existing source network representation pre-trained on large datasets and later fine-tuned using datasets that suit downstream target task, pre-training data bias, pre-trained representation bias and fine-tuning data bias all can be the sources to induce bias in downstream applications. Sentiment analysis \cite{kiritchenko2018examining}, abusive language detection \cite{park2018reducing}, text classification \cite{dixon2018measuring}, machine translation \cite{font2019equalizing,vanmassenhove2018getting}, personalized medicine \cite{rajkomar2018ensuring}, coreference resolution \cite{rudinger2018gender,zhao2018gender}, crime recidivism prediction systems \cite{chouldechova2017fair}, automating resume screening \cite{lambrecht2016algorithmic}, online advertisements delivery \cite{lambrecht2019algorithmic,sweeney2013discrimination}, etc., are some of the downstream applications that reports bias in various domains. Inappropriate or unfair choice of label usages to fine-tune downstream task is another source of bias i.e., \textit{Measurement Bias} \cite{suresh2021framework}.

These biases can be distinguished as \textit{Intrinsic Bias} if it occurs in pre-trained learning or \textit{Extrinsic Bias} if it occurs in downstream task modeling. Besides above mentioned biases, in the perspective of real-world machine learning models, the final system must consider \textit{Evaluation Bias} that occurs when benchmark dataset for a task doesn’t represent certain groups (e.g., images of non-white women not being identified by the model) and \textit{Deployment Bias} that occurs due incompatibility of a model designed for a particular task when used differently (e.g., using risk assessment tool created to predict future crime for the purpose of determining length of sentence) \cite{suresh2021framework}. Table \ref{tab:lm_works} shows a large set of research that identify and mitigate bias from large PLMs, where most of them (around 53\%) focus primarily on bias identification.

\renewcommand{\arraystretch}{1.5}
\begin{longtable}{|l|p{2.35cm}|p{2.1cm}|p{3.7cm}|p{2.7cm}|}
\caption{Works addressing bias in large pre-trained language models}
\label{tab:lm_works}

\\ \hline 
\multicolumn{1}{|c|}{Work} & \multicolumn{1}{c|}{PLM} & \multicolumn{1}{c|}{Domain} & \multicolumn{1}{c|}{Quantification} & \multicolumn{1}{c|}{Mitigation}
\\ \hline 
\endfirsthead

\multicolumn{5}{l}%
{{\tablename\ \thetable{} -- continued from previous page}} 
\\ \hline 
\multicolumn{1}{|c|}{Work} & \multicolumn{1}{c|}{PLM} &
\multicolumn{1}{c|}{Domain} & \multicolumn{1}{c|}{Quantification} & \multicolumn{1}{c|}{Mitigation}
\\ \hline 
\endhead

\hline \multicolumn{5}{r}{{Continued on next page}} \\ %\hline
\endfoot

\hline 
\endlastfoot

\cite{silva2021towards} 
& BERT, XLNet, ALBERT, DistilBERT, RoBERTa, GPT-2 
& Gender 
& WEAT, sequence likelihood, pronoun ranking 
& WEAT scores as an additional loss regularizer\\ 

\cite{ahn2021mitigating} 
& BERT 
& Ethnicity 
& Normalized probability, Categorical bias score 
& M-BERT, Contextual Word Alignment \\

\cite{de2021stereotype} 
& DistilBERT, RoBERTa, XLM-RoBERTa, ALBERT, BERT 
& Gender 
& Skew and stereotype matrices 
& Data augmentation
\\

\cite{fatemi2021improving} 
& BERT 
& Profession, Gender
& Model querying to predict pronouns at masked position, given context 
& GEnder Equality Prompt \\

\cite{bhardwaj2021investigating} 
& BERT 
& Gender 
& MLP regressor 
& Removing word vector components from gender directions \\

\cite{liu2021mitigating} 
& GPT-2 
& Politics 
& Indirect and Direct bias metrics 
& Reinforcement Learning\\

\cite{jin2021transferability} 
& RoBERTa 
& Ethnicity, Gender
& Classification Performance, Group identifier bias metrics, African American Vernacular English, Dialect Bias Metrics, Gender Stereotype Metrics 
& Upstream Bias Mitigation \\

\cite{schick2021self} 
& GPT-2, T5 
& Occupation, Nationality, Religion, Gender, Age, Race, Disability
& Self-Diagnosing 
& Self-Debiasing \\

\cite{abid2021persistent} 
& GPT-3 
& Religion 
& Prompt completion, analogical reasoning, story generation 
& \multicolumn{1}{c|}{---} \\

\cite{stanczak2021quantifying} 
& XLM, XLM-RoBERTa, BERT multilingual 
& Gender  
& Point-wise mutual information and its extension with latent sentiment \& regularization 
& \multicolumn{1}{c|}{---} \\

\cite{kaneko2021unmasking} 
& BERT, RoBERTa, ALBERT 
& Gender, Race, Sexual orientation, Religion, Nationality, Disability,  Age, Physical appearance, Socio-economic status
& All Unmasked Likelihood (AUL), AUL with Attention weights
& \multicolumn{1}{c|}{---} \\

\cite{basta2021extensive} 
& EMLo 
& Gender 
& Gender direction in gender subspace
& \multicolumn{1}{c|}{---} \\

\cite{ye2021adversarial} 
& BERT 
& Gender 
& Prediction probability scores
& \multicolumn{1}{c|}{---} \\

\cite{guo2021detecting} 
& ELMo, GPT, BERT, GPT-2
& Intersectional  
& Intersectional Bias Detection, Emergent Intersectional Bias Detection, CEAT
& \multicolumn{1}{c|}{---} \\

\cite{magee2021intersectional} 
& DistilBERT, GPT-2, GPT-NEO 
& Intersectional bias 
& Difference/similarity between sentiment scores 
& \multicolumn{1}{c|}{---} \\

\cite{wolfe2021low} 
& BERT, GPT-2, T5, XLNet 
& Low frequency names 
& SV-WEAT, Intra and Inter layer self-similarity score 
& \multicolumn{1}{c|}{---} \\ 

\cite{li2021detecting} 
& BERT 
& Gender 
& BERT Attention maps 
& \multicolumn{1}{c|}{---} \\

\cite{kirk2021bias} 
& GPT-2 
& Intersectional 
& Predictions of GPT-2 
& \multicolumn{1}{c|}{---} \\

\cite{nadeem2020stereoset} 
& BERT, RoBERTa, XLNET, GPT2 
& Profession, Gender, Religion, Race 
& Language modelling score, Stereotype score, iCAT 
& \multicolumn{1}{c|}{---} \\

\cite{lu2020gender} 
& LSTM
& Gender 
& Bias Score, Intervention matches 
& Counterfactual Data Augmentation, Word Embedding Debiasing  \\

\cite{zhang2020hurtful} 
& BERT pre-trained on medical notes 
& Ethnicity, Gender, Insurance groups 
& Log probability score, extrinsic evaluation using downstream tasks 
& \multicolumn{1}{c|}{---} \\

\cite{bartl2020unmasking} 
& BERT 
& Profession, Gender
& Association test 
& Name based Counterfactual Data Substitution \\

\cite{vig2020investigating} 
& GPT-2 
& Gender 
& Causal Mediation Analysis 
& \multicolumn{1}{c|}{---} \\

\cite{tan2019assessing} 
& GPT-2, CBoW-GLoVe, ELMo, GPT, BERT 
& Intersectional, Gender, Race
& Counting occurrences, Modified SEAT 
& \multicolumn{1}{c|}{---} \\

\cite{kurita2019measuring} 
& BERT 
& Gender
& Log Probability Bias Score, WEAT
& \multicolumn{1}{c|}{---} \\

\cite{zhao2019gender} 
& ELMo 
& Gender 
& Co-occurrence 
& Train-time data augmentation, test-time neutralization \\

\cite{qian2019reducing} 
& LSTM
& Gender 
& Occurrences, causal testing, Euclidean distance 
& Loss function modification \\

\cite{bordia2019identifying} 
& LSTM
& Gender 
& Fixed \& infinite context bias scores
& Loss function modification \\

\cite{basta2019evaluating} 
& ELMo 
& Gender 
& Direct Bias, Biased words clustering \& classification
& \multicolumn{1}{c|}{---} \\

\cite{huang2019reducing} 
& Transformer-XL
& Country, Occupation, Name 
& Individual \& group fairness metrics from sentiment scores
& Embedding \& Sentiment regularization \\

%\end{tabular}
\end{longtable}

\section{Quantifying Bias in PLMs}
\label{sec:quantify}

In order to study unfavorable consequences of different sources of biases in various domains, it is necessary to quantify bias in some manner. Based on the stages of occurrence, bias are generally quantified in the corpus, representation and downstream tasks.

\subsection{Quantifying bias in corpora}

Counting occurrences or co-occurrences of key terms and deriving various statistics from them helps to discover bias in corpora. Bordia and Bowman \cite{bordia2019identifying} study gender bias in three publicly available datasets, viz., Penn Treebank \cite{marcus1993building}, WikiText-2 \cite{merity2016pointer} and CNN/Daily Mail \cite{hermann2015teaching} that are used to build language models by finding bias scores built using word-level probability profiles within the context of gendered words by defining fixed and infinite sized context windows around the gendered words. Their bias score helps to find whether words more frequently co-occur with female or male gendered words. For fixed sized context windows, they find optimal sized windows as smaller windows can focus more on target words whereas larger windows can focus on the broader topic. Infinite sized windows are much more stable by using exponentially diminishing weights as the distance between key terms and gendered words increases. Tan et al. \cite{tan2019assessing} count occurrences of key terms (e.g. female or male pronouns) and their co-occurrence with stereotypically gendered occupation terms and perform statistical analysis to find gender bias and also racial and intersectional biases in 1 Billion Word Benchmark \cite{chelba2013one}, BookCorpus \cite{zhu2015aligning}, Wikipedia, and WebText \cite{radford2019language} datasets used to pre-train contextual word models.

\subsection{Quantifying bias in representations}

\subsubsection{Geometry of vector spaces}

Certain works quantify bias by analyzing subspaces in embeddings. Bolukbasi et al. \cite{bolukbasi2016man} demonstrates the occurrence of gender bias in word embeddings by viewing the difference between word vectors of gendered words like \textit{`sister'}, \textit{`brother'}, \textit{`grandmother'}, \textit{`grandfather'}, etc., and gender neutral words. They observe a direction that essentially captures gender and projecting gender neutral words in this direction helps them to quantify gender bias. Manzini et al. \cite{manzini2019black} extends this approach to suit multiclass settings and demonstrate that word representations exhibit several stereotypical biases including religion based and racial bias. 

\subsubsection{Word association test}

Word association tests to quantify representation bias are inspired from the Implicit Association Tests (IAT) in psychology \cite{greenwald1998measuring} that tries to understand human subconscious bias by measuring differences in the association of target concepts with an attribute. Caliskan et al. \cite{caliskan2017semantics} proposes a statistical test called the Word Embedding Association Test (WEAT) analogous to IAT to quantify bias in non-contextual word embeddings GloVe and Word2Vec. Using WEAT, authors examine the similarity of embeddings of words in complementary categories like European American and African American names with the complementary attributes like pleasant and unpleasant attributes. The dissimilarity between the association of European American names with the attributes when compared to African American names with the same attributes, helps their study to report the existence of human-like implicit bias in embeddings. To test bias in sentence encoders like ELMo and BERT, May et al. \cite{may2019measuring} proposes a generalization of WEAT named the Sentence Encoder Association Test (SEAT). Using SEAT, even though they could verify presence of bias in these embeddings, the results were not very generalizable. They also point out that dissimilarities in the results don't mean contextual embeddings is free of bias, rather it may be an indication of cosine similarity not suitable to measure the similarity of embeddings of recent contextual models, and hence an alternative might be required to quantify bias in such representations.  Kurita et al. \cite{kurita2019measuring} also shows that conventional cosine similarity based methods don't produce consistent results to find bias in sentence embeddings generated from the contextual models, as the embeddings of the words may differ according to the context and state of the language model. As the former tests concentrate only on individual words and predefined stereotypical attributes like pleasant and unpleasant terms in artificial contexts that don't reflect the natural use of words, Nadeem et al. \cite{nadeem2020stereoset} proposes two Context Association Tests (CAT) to intrinsically estimate bias in a set of large PLMs. They perform CAT at the sentence level and discourse level, where each target term has natural context, and observe that the recent contextual models BERT, GPT2, RoBERTa, and XLNet exhibit strong biases with respect to their language modeling ability. Building upon WEAT, Guo and Caliskan \cite{guo2021detecting} propose Contextualized Embedding Association Test (CEAT) to confirm and extensively quantify biases in neural language models ELMo, BERT, GPT and GPT-2 according to different contexts.

\subsection{Quantifying bias in downstream tasks}

In this category, bias is quantified by checking performance scores of system over evaluation corpus that differs only in the context of target terms in which domain of bias is being studied. For example, gender swapping to change gender of gendered words, like \textit{`She is here’} to \textit{`He is here’}, and then evaluating model performance of these two sentences \cite{kiritchenko2018examining,lu2020gender,rudinger2018gender,zhao2018gender}. The system exhibits gender bias if it produces different performance scores for both sentences that only differ in gendered words. Dixon et al. \cite{dixon2018measuring} presents two performance evaluation metrics derived from error rate equality difference to quantify bias in text classifier constructed to identify toxic comments. Park et al. \cite{park2018reducing} utilizes these metrics to quantify bias along with the method to generate gender unbiased dataset proposed by Dixon et al. \cite{dixon2018measuring} to find gender bias in abusive language detection. Kiritchenko and Mohammad \cite{kiritchenko2018examining} utilize difference in predicted intensity scores of sentences that differ in gendered words and race, over their corpus to statistically evaluate gender and racial bias in 219 sentiment analysis models that took part in {\it SemEval 2018: Task 1 Affect In Tweets}. Zhao et al. \cite{zhao2018gender} illustrates gender bias in coreference systems using F1 score over their evaluation corpus named WinoBias that contains pair of sentences that associate gendered pronouns (her/him) to various female or male stereotypical occupations (secretary/physician). Rudinger et al. \cite{rudinger2018gender} also perform a similar gender bias study in coreference system over pairs of sentences that differ only in gendered words. Lu et al. \cite{lu2020gender} quantifies gender bias in coreference resolution and language modeling by calculating dissimilarity in performances of gendered words with various occupations across pairs of sentences that are gender swapped. In addition, bias can also be measured using interpretability, by investigating model interpretations on how it reaches to certain decisions or predictions \cite{du2020fairness}. 

\section{Mitigating Bias in PLMs}
\label{sec:mitigating}

Various efforts have been made for mitigating bias in PLMs to reduce or remove their discriminatory influences over underrepresented or non-mainstream groups. We classify debiasing approaches into three categories, based on the stages at which they are addressed. 

\subsection{Mitigating bias in training corpora}

Most computational algorithms designed for extrinsic applications like sentiment analysis, text classification, etc., primarily rely on labeled training corpora making them prone to societal biases present in these corpora \cite{torralba2011unbiased}. Despite this being the case, such corpora are still being extensively utilized in various applications of NLP as it is expensive or labor-intensive to build new large-sized training corpora. Hence, generally, common techniques such as data augmentation and bias fine-tuning are followed for debiasing training corpora.

\subsubsection{Data augmentation}

Data augmentation techniques debias the training corpus by supplying additional data to support the target groups with comparatively fewer data in the corpora and thereby creating a balanced corpus to train on. Augmentation can thus in a way counterbalance the under/over representations of any particular target group of a domain on training corpus. Zhao et al. \cite{zhao2019gender,zhao2018gender} adopts data augmentation to reduce gender bias in the downstream task of coreference resolution while observing that the possibility of associating occupations to masculine terms is very much higher than feminine terms. They augment gender swapped versions of data, from masculine terms to feminine terms and vice-versa, to the training corpus after anonymizing the named entities. Park et al. \cite{park2018reducing} utilizes the idea of data augmentation proposed by Zhao et al. \cite{zhao2018gender} to reduce gender bias in abusive language detection. Very similar to Zhao et al. \cite{zhao2018gender}, Lu et al. \cite{lu2020gender} proposes Counterfactual Data Augmentation to explore gender bias in Neural NLP applications including coreference resolution and language modeling by representing bias with reference to internal scores in neural models. Apart from the attempts to remove gender bias in English language, Zmigrod et al. \cite{zmigrod2019counterfactual} proposes variation to naive gender swapping based counterfactual data augmentation to manage gender bias in morphology rich or inflected languages like Spanish and Hebrew that otherwise deliver ungrammatical sentences with simpler approaches. Maudslay et al. \cite{maudslay2019s} improves counterfactual data augmentation by proposing two variants named Counterfactual Data Substitution and Names Intervention to address indirect bias. Liu et al. \cite{liu2019does} use concept of Maudslay et al. \cite{maudslay2019s} and propose Counterpart Data Augmentation to remove biases in dialogue systems. Even though data augmentation based debiasing is a simple technique to reduce bias in the training corpora, their annotation is expensive and increases the size of dataset, in turn, increasing the time required for training. Moreover, these techniques generally only consider isolated words to perform binary swapping and mostly ignore non-binary and more sophisticated representations in a domain. Also, these techniques rely on a pre-defined limited list of key terms and associated pairs, which may be conceivably incomplete, where some terms may have different spelling (e.g. \textit{mommy} vs. \textit{mummy}), different morphology (e.g. \textit{his}\textrightarrow \textit{her} and \textit{his}\textrightarrow \textit{hers}), pairing variations (e.g. \textit{breastfeed}), or produce absurd sentences due to blind swaps (e.g. \textit{she is pregnant}\textrightarrow \textit{he is pregnant}).

\subsubsection{Bias fine-tuning}

Another approach to debias training corpora is bias fine-tuning by transfer learning, where, instead of the expensive process of constructing balanced corpora with respect to a domain, the idea of transfer learning is employed. As data biases generally originate from data imbalance or small sized datasets, to make sure the model doesn't overfit on biased data, Park et al. \cite{park2018reducing} utilize transfer learning to gender debias abusive language detection model by regularizing from a source network trained on a less biased large dataset and thereafter fine-tune on a target dataset that is largely gender biased. Even though the approach significantly reduces bias, it seems to hurt the overall model accuracy too (as may be expected). Saunders et al. \cite{saunders2020reducing} proposes an approach almost opposite to Park et al. \cite{park2018reducing} where they execute small-domain adaptation by fine-tuning on a less biased small dataset to address gender bias in neural machine translations.

\subsection{Mitigating bias in representation}

An earlier approach to debias the representations is geometric debiasing that removes subspace of protected domain/target terms concerned to a domain in embedding. For a domain, this post-process approach identifies subspace or direction in embedding that holds bias and removes the association of neutral words to target words concerning that domain. This hard debiasing approach is followed by Bolukbasi et al. \cite{bolukbasi2016man} to alleviate gender bias from word embeddings by levelling the distance of gender neutral words towards the set of gendered words. They also propose soft bias correction that aims to conserve the initial embedding distances using a trade-off parameter to balance with debiasing. But as it works by removing gender information from the words, it might not be a generalizable approach such as in social science and medical applications that makes use of these gender information \cite{back2010gender,mcfadden1992study}. Also, their approach make use of classifier to distinguish gender neutral words that in turn can propagate classification errors, influencing the performance of debiasing \cite{zhao2018learning}. Zhao et al. \cite{zhao2018learning} neutralize embedding with respect to gender by locating gender neutral words along with the process of training word vectors without employing an additional classifier. However, Gonen and Goldberg's \cite{gonen2019lipstick} experiments shows that the approaches of Bolukbasi et al. \cite{bolukbasi2016man} and Zhao et al. \cite{zhao2018learning} are insufficient blind debiasing techniques that just hides but doesn't actually remove bias. For word level language models, Bordia and Bowman \cite{bordia2019identifying} make use of a loss function regularizer to penalize the projection of embeddings onto gender subspace as a soft debaising version of Bolukbasi et al. \cite{bolukbasi2016man}. As stated by Gonen and Goldberg \cite{gonen2019lipstick}, they also mention chances of bias even after debiasing the representations, as a case that their bias score may be not able to detect it. Alternative to this direct geometric debiasing, Zhao et al. \cite{zhao2019gender} proposes a different gender neutralisation procedure by averaging the representations of original and corresponding gender swapped data versions for debiasing contextualised representations. To mitigate sentiment bias in word embedding concerned to demographic domains, Sweeney et al. \cite{sweeney2020reducing} removes the correlations of demographic terms with sentiments. 

\subsection{Mitigating bias in algorithm}

Debiasing representations are harder to be applied in contextual embeddings, since the representation of a word can vary according to the plurality of different contexts it appears in \cite{magee2021intersectional}. In such cases, certain works focus at modifying algorithm/model, i.e., pre-trained or fine-tuned models, in such a way that it mitigates the bias in predictions during training (i.e., in-processing), usually by modifying the loss function of the model. Qian et al. \cite{qian2019reducing} alters the loss function of the language model in such a way that the model adapts to equalize prediction probabilities for gendered pairs of words thereby reducing output gender bias.  Silva et al. \cite{silva2021towards} consider utilising WEAT for debasing transformer based PLMs and later use WEAT score along with cross entropy to modify loss function of RoBERTa. Huang et al. \cite{huang2019reducing} propose another variation of altering regular cross entropy loss function with embedding and sentiment driven regularisation terms for mitigating sentiment bias in large PLMs. Adversarial training is another way to debias algorithms by altering loss functions while training a predictor along with an adversary, which is intended to produce fair prediction by minimising the adversarial function that tries to model protected domains like gender, race, etc. \cite{zhang2018mitigating}. A similar approach is utilised by Zhang et al. \cite{zhang2020hurtful} to mitigate gender bias in clinical contextual word embeddings. A different recent approach proposed by Liu et al. \cite{liu2021mitigating} uses reinforcement learning to debias political bias in large PLMs by utilising rewards from word embedding or a classifier. 

\section{Affective Bias}
\label{sec:affective_bias}

Textual affective computing involves development of algorithms that accurately identify what is subjectively written in text and how to make better affect-influenced subjective decisions, i.e., decisions that are based on emotions, sentiments, or opinions, in several real-world systems. Many systems were initially proposed to detect and measure affect (emotion or sentiment) expressed by textual data such as in movie reviews, product reviews, etc. \cite{pang2002thumbs,suharshala2018cross}. Later these systems were widely adopted into various domains such as healthcare \cite{gupta2016twitter}, commercial applications \cite{krishnamoorthy2018sentiment,renault2020sentiment,suharshala2018cross}, politics \cite{caetano2018using}, education \cite{dolianiti2018sentiment,suharshala2018cross} and many more. When industrial giants like Google\footnote{\url{ https://cloud.google.com/natural-language/docs/analyzing-sentiment}}, IBM\footnote{\url{https://cloud.ibm.com/apidocs/natural-language-understanding\#emotion}} and Microsoft\footnote{\url{https://docs.microsoft.com/en-us/azure/cognitive-services/language-service/sentiment-opinion-mining/overview}} developed natural language understanding tools, textual affective understanding became a crucial and significant part of them. Several observations and arguments have been made by researchers to demonstrate the importance of textual affective computing tasks, especially towards sentiment analysis. According to Cambria et al. \cite{cambria2017sentiment} sentiment analysis task still needs to travel much to reach human level performance which can be attained by successfully solving several other NLP problems such as POS tagging, Named Entity Recognition, etc. Poria et al. \cite{poria2020beneath} draw attention to optimistic future research directions in affecting computing like multi-modal affective computing, sarcasm analysis and even the contemporary issue of social bias in affective NLP systems, and strive to oppose the conventional belief that sentiment analysis tasks are saturated being 20 years old. Our study on affective bias in NLP is motivated from these observations on heightened relevance of affective computing and its wide applicability in diversified NLP applications that leverage affect information \cite{anoop2020affect,anoop2020emotion,elmadany2020leveraging}.     

\subsection{Definition and implications}

The study on bias in NLP and machine learning was heavily accelerated through observations on the negative impact of NLP biases towards certain social or marginalized groups when applied in the real world\footnote{\url{https://www.propublica.org/article/machine-bias-risk-assessments-in-criminal-sentencing }}. Affective bias is another recent research category in this direction, where researchers study any unfair or imbalanced associations of affect with key terms representing particular underrepresented or protected groups in a domain and how it influences NLP systems, such as sentiment and emotion detection. For example, Google sentiment analyzer is observed to infer that \textit{being gay}\footnote{\url{https://www.vice.com/en/article/j5jmj8/google-artificial-intelligence-bias}} is bad and assigns high negative sentiments towards sentences such as \textit{`I’m a gay black woman’}  and \textit{`I’m a homosexual'}. By the term Affective bias in NLP, we define the existence of unfair or biased associations of affect (maybe emotions such as anger, fear, joy, etc. or sentiment such as positive, neutral, negative) towards underrepresented groups, or over-generalized beliefs (stereotypes) about particular social groups in textual documents. For example, in textual documents, words associated with women such as \textit{`she’}, `\textit{wife’}, etc., are highly associated with a certain category of emotions like \textit{`sadness’} and \textit{`anger’}, and representations of the Muslim religion are observed as being associated with negative terms that indicate violence \cite{abid2021large}, etc. The existence of affective bias in textual affective computing systems harms its utility and applicability in tasks like business and commercial decision making, healthcare, etc. The concept of affective bias is valid and applicable beyond the NLP frameworks because, as in \cite{buolamwini2018gender}, there are chances of high classification error rates for facial emotion detection systems towards underrepresented social groups. Similarly, it is crucial to evaluate all kinds of affective computing systems in the backdrop of affective bias, since automated emotion detection systems have a huge impact on modeling human behavior in many intelligent artificial artifacts or algorithms that imitate human emotion systems for their completeness.    

\subsection{Affective bias in PLMs}

Many textual affective computing systems including lexicon based \cite{rozado2020wide}, conventional machine learning \cite{sweeney2020reducing}, deep learning \cite{bhaskaran2019good,shen2018darling}, and hybrid \cite{diaz2018addressing} approaches perpetuate affective bias, which, in general, is transmitted from emotional bias of humans through models learnt over large scale textual corpora. For example, usage of a corpus that contains textual data where humans had expressed the \textit{anger} emotion towards a particular underrepresented religion for training an algorithm can later propagate or/and amplify this affect oriented bias stereotyping religion with emotion \textit{anger}. Table \ref{tab: affective_bias} illustrates an extensive snapshot of works in this area of research along with their major characteristics. A predominant part of existing works study affective bias specific to gender and through the perspective of sentiment analysis \cite{bhaskaran2019good,kiritchenko2018examining,rozado2020wide,shen2018darling,sweeney2020reducing}. Whereas, other domains like religion, politics, intersectional biases, etc., and their impact in perspective of fine-grained emotion classes (anger, fear, joy sadness, surprise, disgust) have not been investigated as much, except in \cite{kiritchenko2018examining} and \cite{venkit2021identification}. The inadequacy of generic evaluation corpora to perform affective bias evaluation in various domains along with the evaluation sentences having corresponding emotion/sentiment ground truth or output label is a notable gap in this area of research. When most works in literature try to identify the existence of affective bias in NLP systems, only very few explore mitigation of the harms of affective bias. The body of work in affective bias can be split into two categories, conventional approaches and runtime verification approaches. Conventional way of analyzing affective bias tries to identify bias in training data, algorithm, representation, fine-tuning data, and finally, mitigates them using different strategies \cite{kiritchenko2018examining,sweeney2020reducing,venkit2021identification}. Whereas, runtime verification approaches monitor and uncover biased predictions in each run of a specified system using mutations of input sentences (automatically generated templates from input sentences and their paired sentences that represent different views of a stereotype). Runtime verification is generally suitable to validate whether the system satisfies fairness criteria over each run. 

The conventional approach by Shen et al. \cite{shen2018darling} investigates bias in sentiment prediction for textual write-ups comprising similar content generated by different groups of people. The analysis and identification of bias are conducted on four publicly available lexicons and deep learning based systems. A similar approach by Zhiltsova et al. \cite{zhiltsova2019mitigation} also identifies and mitigates sentiment bias against non-native English text by using four popular lexicon based emotion prediction systems. Both these works rely on linguistic style changes across different human groups and how it leads to affective bias in NLP. Apart from analysis of affective bias in lexicon and conventional machine learning systems, researchers also explore non-contextual word embeddings such as word2vec, GloVe, and FastText in the context of affective bias \cite{diaz2018addressing,rozado2020wide,sweeney2020reducing}. A significant contribution in this regard is the work by Diaz et al. \cite{diaz2018addressing} addressing age related affective bias in ten widely used word embeddings and fifteen different sentiment analysis models. The work primarily validates whether opinion polling systems falsely report any age group (old or young) more negatively or positively, for example a sentence with adjectives of \textit{`young’} more likely scores positive sentiments than the same sentence with adjectives of \textit{`old’} \cite{diaz2018addressing}. Among the similar studies based on non-contextualised word embeddings, Sweeney et al. \cite{sweeney2020reducing} introduce an adversarial learning strategy to mitigate demographic affective bias in word2vec and GloVe, and Rozado et al. \cite{rozado2020wide} screen word embeddings to identify bias through the notion of representing words along cultural axis in the embedding space. A more generic work of Kiritchenko et al. \cite{kiritchenko2018examining}  identify affective bias in two hundred emotion prediction systems that participated in the shared task {\it SemEval-2018 Task 1 Affect in Tweets}. They procure an evaluation corpus, Equity Evaluation Corpus (EEC), one of the few publicly available evaluation corpus that has generic evaluation sentences and ground truth emotion labels as basic emotions anger, fear, joy, and sadness for all evaluation sentences in corpus. A similar evaluation corpus has been proposed by Venkit et al. \cite{venkit2021identification} considering sentences from the domain of persons with disabilities.

A more noteworthy approach to quantify affective bias concerning occupational stereotypes in contextualized large PLM, BERT, is discussed in \cite{bhaskaran2019good}. Even though more works recently identify and mitigate generic bias in large PLMs due to their efficacy and utility \cite{liang2021towards,nadeem2020stereoset}, very few of them investigate affective bias in large PLMs. Notable works to uncover bias in sentiment analysis systems that utilize popular PLMs such as Google BERT, Facebook RoBERTa, Google ALBERT, Google ELECTRA, and Facebook Muppet rely on runtime verification approach instead of conventional paradigms to analyze bias \cite{asyrofi2021biasfinder,yang2021biasrv}. Another interesting approach by Huang et al., \cite{huang2019reducing} investigates sentiment bias introduced in text generated by language models. This emerging scenario facilitates to conduct more evaluations to identify and mitigate affective bias in large PLMs such as BERT, ALBERT, RoBERTa, XLNet, GPT, etc.

\renewcommand{\arraystretch}{1.5}
\begin{longtable}{|l|p{2.55cm}|p{3.6cm}|p{1.8cm}|p{2.9cm}|}
\caption{Works addressing affective bias}
\label{tab: affective_bias}

\\ \hline 
\multicolumn{1}{|c|}{Work} & \multicolumn{1}{c|}{Domain} & \multicolumn{1}{c|}{Quantification} & \multicolumn{1}{c|}{Mitigation} & \multicolumn{1}{c|}{Model}
\\ \hline 
\endfirsthead

\multicolumn{5}{l}%
{{\tablename\ \thetable{} -- continued from previous page}} 
\\ \hline 
\multicolumn{1}{|c|}{Work} & \multicolumn{1}{c|}{Domain} & \multicolumn{1}{c|}{Quantification} & \multicolumn{1}{c|}{Mitigation} & \multicolumn{1}{c|}{Model}
\\ \hline 
\endhead

\hline \multicolumn{5}{r}{{Continued on next page}} \\ %\hline
\endfoot

\hline 
\endlastfoot

\multicolumn{5}{|c|}{Sentiment perspective} \\
\hline

\cite{yang2021biasrv} 
& Gender 
& BiasFinder in \cite{asyrofi2021biasfinder}
& \multicolumn{1}{c|}{---} 
&  BERT \\

\cite{asyrofi2021biasfinder} 
& Occupation, Country of origin, Gender
& Metamorphic Testing 
& \multicolumn{1}{c|}{---} 
& BERT, RoBERTa, ALBERT, ELECTRA, Muppet \\

\cite{sweeney2020reducing}
& Gender
& Directional sentiment vectors 
& Adversarial learning
& SVM, LSTM \\

\cite{rozado2020wide}
& Ethnicity, Age, Sociodemographic status, Physical appearance, Religion, Politics, Gender
& Projecting word embeddings to cultural axis
& \multicolumn{1}{c|}{---} 
& Lexicon based \\

\cite{bhaskaran2019good}
& Occupation, Gender
& Statistical significance difference
& \multicolumn{1}{c|}{---} 
& BERT, Bi-LSTM, Logistic Regression \\

\cite{diaz2018addressing}
& Age 
& Multinomial log-liner regression, paired t-test
& \multicolumn{1}{c|}{---}
& Lexicon based, conventional machine learning, hybrid \\

\cite{shen2018darling}
& Gender, Race, Politics
& Difference in mean sentiment score, statistical significance test, linear regression 
& \multicolumn{1}{c|}{---} 
& Rule based, Naive Bayes, Dynamic CNN, LSTM\\

\hline
\multicolumn{5}{|c|}{Emotion perspective} \\
\hline

\cite{venkit2021identification}
& Gender, Race 
& Mean score of prediction, linear regression on sentiment scores
& \multicolumn{1}{c|}{---} 
& DistilBERT, TextBlob, Google API, VADER \\

\cite{zhiltsova2019mitigation}
& Non-native English speaker
& Wilcoxon signed rank test
& Lexical score 
& VADER, Afinn, SentimentR, TextBlob \\

\cite{kiritchenko2018examining}
& Gender, Race
& Average score difference
& \multicolumn{1}{c|}{---} 
& Deep Learning, conventional machine learning, lexicon based
%\end{tabular}
\end{longtable}

\section{Bias Evaluation Corpora}
\label{sec:evaluation_corpora}

Ideal bias evaluation corpora are indispensable components that helps to identify and measure the existence of different types of bias in NLP systems. Since this review focuses on bias in PLMs with special emphasis on affective bias, we present popular and publicly available bias evaluation corpora along with their limitations in the backdrop of PLMs and affective biases, in table \ref{tab:evalutaion_corpora}; some of the evaluation corpora are avoided due to their unavailability in the public domain to aid future research  (e.g., \cite{huang2019reducing,zhiltsova2019mitigation}). The most common attempt to create bias evaluation corpora are by initially building template sentences and interchanging its key terms (e.g. \textit{she}\textrightarrow \textit{he}, \textit{wife}\textrightarrow \textit{husband}, \textit{mother}\textrightarrow \textit{father}) associated to a target (e.g. Female, Male) concerning a domain (e.g. Gender) which induces bias in natural language.  For example, Diaz et al. \cite{diaz2018addressing} creates bias evaluation corpora intended to identify age related sentiment bias using template sentences like  \textit{`This \textlangle AGE RELATED KEY TERM\textrangle guy was 3 or 4 feet from the tide line and the tide was going out'} and interchanging \textlangle AGE RELATED KEY TERM\textrangle with \textit{`old'} and \textit{`young'}. Contrary to conventional \textit{interchanging of key terms} approach, StereoSet \cite{nadeem2020stereoset} is a much diverse, large-scale and natural bias evaluation corpus procured through inter and intra sentence Context Association Test using Amazon Mechanical Turk.  Design and creation of such bias evaluation corpora are often defined carefully to isolate and extract the real effect of bias in various stages such as data bias, algorithm bias, etc., and also to avoid presence of any biased data within evaluation corpus or unbalanced set of key terms referring to a particular target in a domain. 

Evaluating large PLMs has recently become highly essential particularly due to their increasing usage in many real-world applications. Several existing bias evaluation corpora are useful enough to evaluate large PLMs when considering the total number, i.e., size, of available evaluation sentences in these corpora \cite{dixon2018measuring,kiritchenko2018examining,zhao2018gender}. But these corpora fail to represent diverse real-world contexts as they are generally built synthetically using very simple and short-length sentences that are far from real-world scenarios and may even deliver nonsensical samples (e.g. \textit{`my \textlangle sister\textrangle is pregnant'}\textrightarrow \textit{`my \textlangle brother\textrangle is pregnant'}).  To accurately design an efficient bias evaluation corpora that suit evaluating PLMs, Liang et al. \cite{liang2021towards} leverage naturally occurring text corpora from WIKITEXT-2 \cite{merity2016pointer}, SST \cite{socher2013recursive}, etc., to generate evaluation sentences that can represent real-world context better than the corpus in \cite{dixon2018measuring,kiritchenko2018examining,nadeem2020stereoset,sheng2019woman,zhao2018gender}. There is also a need for more efficient evaluation corpora to evaluate affective bias; one of the available corpus by Kiritchenko et al. \cite{kiritchenko2018examining} contains, as mentioned earlier, very simple synthetic and far from real-world templates where emotions are specified in the sentence explicitly, e.g., \textit{`My brother made me feel angry’} with ground-truth label \textit{anger}. Another affect oriented corpus by Venkit et al. \cite{venkit2021identification} much suitably represents real-world context suitable to evaluate PLMs, however, they are specifically designed to address bias concerned to people with disabilities, e.g.\textit{ `They were aggravated because of the Mentally Handicapped neighbour’} with ground-truth label \textit{anger}. To effectively test and build fair machine learning software, it is highly essential to develop more generic and real-world context enabled evaluation corpora addressing various domains like religion, race, intersectional bias, non-binary representations of gender (including lesbian, gay, etc., apart from common way of only addressing gender as a binary domain). Such corpora can eventually lead to the usage of bias evaluation corpora as an integral part of any NLP system to evaluate system bias like any other test performed on the system before being deployed in real-world \cite{chakraborty2020fairway}. 
\renewcommand{\arraystretch}{1.5}
\begin{longtable}{|p{4.5cm}|p{1.6cm}|p{1.8cm}|p{3.9cm}|}
\caption{Bias evaluation corpora}
\label{tab:evalutaion_corpora}
%\begin{tabular}{|l|p{1.9cm}|p{1.5cm}|p{1.8cm}|p{2.5cm}|p{2.5cm}|}
% \hline
%Reference &  Large PLM & Domain & Occurence & Quantification & Mitigation\\
%\hline

\\ \hline 
\multicolumn{1}{|c|}{Corpora} & \multicolumn{1}{c|}{Domain} & \multicolumn{1}{c|}{Model} & \multicolumn{1}{c|}{Limitation}
\\ \hline 
\endfirsthead

\multicolumn{4}{l}%
{{\tablename\ \thetable{} -- continued from previous page}} 
\\ \hline 
\multicolumn{1}{|c|}{Corpora} & \multicolumn{1}{c|}{Domain} & \multicolumn{1}{c|}{Model} & \multicolumn{1}{c|}{Limitation}
\\ \hline 
\endhead

\hline \multicolumn{4}{r}{{Continued on next page}} \\ %\hline
\endfoot

\hline 
\endlastfoot

BITS \cite{venkit2021identification} \url{https://github.com/PranavNV/BITS}
& Gender, Race 
& Emotion detection
& In context of people with disabilities, not generic \\

\cite{liang2021towards} \url{https://github.com/pliang279/LM_bias}
& Gender, Religion
& Large PLM
& \multicolumn{1}{c|}{---}  \\

StereoSet \cite{nadeem2020stereoset} \url{https://github.com/moinnadeem/StereoSet}
& Gender, Race, Religion, Profession
& Large PLM
& Less diverse to represent real-world contexts\\

BEC-Pro \cite{bartl2020unmasking} \url{https://github.com/marionbartl/gender-bias-BERT}
& Gender, Profession
& Large PLM
& Less diverse to represent real-world contexts\\

GenderCorpus \cite{bhaskaran2019good} \url{https://github.com/jayadevbhaskaran/gendered-sentiment}
& Gender, Profession
& Sentiment analysis 
& Groud-truth sentiment labels are not available \\

AgeBias \cite{diaz2018addressing} \url{https://dataverse.harvard.edu/dataset.xhtml?persistentId=doi:10.7910/DVN/F6EMTS}
& Age
& Sentiment analysis
& Groud-truth sentiment labels are not available \\

EEC \cite{kiritchenko2018examining}  \url{https://saifmohammad.com/WebPages/Biases-SA.html}
& Race, Gender
& Emotion detection
& Explicit representation of emotions, small \& simple sentences, less diverse to represent real-world contexts\\

Winograd \cite{rudinger2018gender} \url{https://github.com/rudinger/winogender-schemas}
& Gender, Profession
& Coreference resolution
& Small sentences\\

WinoBias \cite{zhao2018gender} \url{https://github.com/uclanlp/corefBias/tree/master/WinoBias/wino}
& Gender, Profession
& Coreference resolution
& Miss ambiguous pronouns in sufficient volume or diversity to accurately represent practical utility of models\\

GAP \cite{webster2018mind}  \url{https://github.com/google-research-datasets/gap-coreference}
& Gender
& Coreference resolution
& \multicolumn{1}{c|}{---}  \\

Identity synthetic dataset \cite{dixon2018measuring}  \url{https://github.com/conversationai/unintended-ml-bias-analysis}
& Human identity (white, gay, etc.)
& Text classification
& Small and simple sentences, less diverse to represent real-world contexts

%\end{tabular}
\end{longtable}

\section{Discussion on Research Challenges}
\label{sec:research_challenges}

Here we discuss several challenges in NLP bias, specifically in the context of large PLMs and also in the context of affective bias.

\subsubsection{Heterogeneous nature of NLP bias}
A large number of research approaches to identify and mitigate NLP biases generally address gender bias (as can be observed from table \ref{tab:domain}), even though there exist many other domains such as race, religion, intersectional biases, etc., that needs to be addressed significantly. Tan et al. \cite{tan2019assessing} provides evidence in this context that racial bias is encoded strongly in contextualized large PLMs, probably even more than gender bias. They also show that the less explored domain of intersectional biases that consider a mixture of two or more domains (e.g. African American Females) is even more higher than primary biases. Also, most existing research only concentrate on addressing a subset of target terms concerned to a domain, e.g. considering gender domain with binary targets terms male and female, instead of its non-binary nature comprising of other target terms like gay and lesbian. All these observations illustrate that the heterogeneous nature of bias in different perspectives must be considered in future works of evaluating bias to make the NLP systems fully debiased. 

\subsubsection{Evaluation corpus}
Many existing bias evaluation corpora demonstrate the scarcity of realism in sentences, such as synthetic sentences that are poor in representing real-world context, as well as short-length sentences, etc. Accurate understanding and quantification of existing bias in NLP models may not be effective with these corpora, especially in case of large PLMs, which are capable of generating complex real-world sentences to produce benchmark results in many downstream applications. Most of the large real-world evaluation corpora focus on gender domain, and real-world context based corpora that suit other domains are scarce. In case of quantification and mitigation of affective biases, many researchers use generic evaluation corpora in \cite{bhaskaran2019good,diaz2018addressing} where evaluations do not have an affective output label. A corpus developed by Kiritchenko et al. \cite{kiritchenko2018examining} contains relevant emotion labels but lacks real-world contexts and has labels largely derived from explicit mentions of emotions. 

\subsubsection{Generalizable design} Generalizability of any system enables it to be used directly or indirectly within other allied systems. Bias evaluation is becoming an essential part in NLP systems to produce fair decisions by avoiding serious societal harms. This necessity can be achieved largely by designing generalizable bias identification and mitigation modules. Generalizability in the context of evaluation corpora, metrics to identify and quantify the impact of bias, performance trade-off measures post debiasing, are important challenges to be addressed in the future to bring the vital bias evaluation process as a common and simple component that any NLP system can adapt and experiment.

\subsubsection{Pre-train vs. Fine-tune bias} Bias in downstream applications that utilize pre-trained models originate across different stages of system development. Initially, it may be from massive amount of data used to pre-train the model which may get amplified by the pre-training algorithm. The other way of introducing bias is through fine-tuning data as well as fine-tuning algorithm. An existing challenge in this background is to analyze and understand which part of the overall system really causes bias, (i.e., is it from pre-trained or task specific fine-tuning modules), how bias is perpetuated with the harmonized workflow of system components, which parts of overall system must undergo a mitigation process, etc. 

\subsubsection{Linguistic diversity and NLP bias} Recent NLP research, widely explores the utility of developing large PLMs and downstream applications in various different languages by including capability to handle monolingual as well as bilingual strategies. Transformer based large PLM, IndicBERT\footnote{\url{https://huggingface.co/ai4bharat/indic-bert}, accessed on January 03 2022} which handles low resource and morphologically highly inflected Indian languages like Malayalam, is an example. Similarly, a large variety of monolingual and bilingual corpora\footnote{\url{https://huggingface.co/datasets}, accessed on January 03 2022} (in languages English, Afrikaans, Arabic, etc.) and machine learning models\footnote{\url{https://huggingface.co/models}, accessed on January 03 2022} (for various tasks including question answering, text classification, etc.) in different languages illustrate same trends. In this context, it is highly essential to study bias in such different linguistic corpora and the mono and multilingual PLMs learned from those corpora. Among the very few attempts in this direction is the research to mitigate gender bias in Hindi word embedding \cite{pujari2019debiasing}, evaluating gender bias in Hindi-English machine translation \cite{ramesh2021evaluating}, addressing gender bias in languages like Spanish and Hebrew \cite{zmigrod2019counterfactual}, etc. 

\subsubsection{Interdisciplinary methods to analyze NLP bias} 
Sun et al. \cite{sun2019mitigating} states that many techniques deployed in non NLP systems can be applied directly or with small modifications to identify and mitigate bias in NLP. Several studies indicate that bias in NLP systems opens up scope to associate with other branches such as sociology, psychology, socio-linguistics, engineering, legal studies, etc.  The Implicit Association Test \cite{greenwald1998measuring} in psychology and its diverse computational representations such as Word Embedding Association Test \cite{caliskan2017semantics} and Sentence Encoder Association Test \cite{may2019measuring} to quantify representation bias are examples to showcase interdisciplinary nature of removing bias in NLP. Also, lessons from Software Engineering\footnote{``IndoML-2021- Day 1 Session 1'', YouTube video, 01:12:50, Posted by ``CSE IIT Gandhinagar'' December, 16, 2021, \url{https://www.youtube.com/watch?v=y3t8pc1s0Yw}, accessed on January 03 2022} such as unit testing, behavior testing, etc., can be adopted to evaluate and quantify bias in machine learning models at different levels \cite{ribeiro2020beyond}. Sun et al. \cite{sun2019mitigating} treats mitigation of NLP bias as a combined problem of sociology and engineering, where sociology can identify how really humans perceive and encode social bias into language. Research towards interdisciplinary discussions can bring light into current bias quantification and mitigation strategies and inspire developing advanced and practically relevant approaches \cite{avin2015homophily,beukeboom2019stereotypes,schluter2018glass}.

\section{Conclusion}
\label{sec:conclusion}

In this survey, we conducted a comprehensive investigation towards bias in large pre-trained language models, especially the transformer based models. We discuss different types of biases that originate at various stages of pre-trained language model workflow and the methods used to quantify and mitigate these biases. Due to the widespread utility of affective computing systems in the real-world, we give a special emphasis on the less explored area of bias that associates with affect, i.e., affective bias. Our study also lists the popular and publicly available evaluation corpora that aid future research, along with their suitability in large pre-trained language models. Finally, we discuss the challenges in this area of research, addressing which would aid the community to further improve the tasks of identification and mitigation of bias in pre-trained language modes. Materials regarding this survey will be made publicly available at \url{https://github.com/anoopkdcs/NLPBias} and we will keep updating it to aid future research.  

% ---- Bibliography ----
%
% BibTeX users should specify bibliography style 'splncs04'.
% References will then be sorted and formatted in the correct style.
%
\bibliographystyle{splncs04}
\bibliography{reference}
\end{document}